\documentclass[10pt, conference]{IEEEtran}
\IEEEoverridecommandlockouts

\usepackage{booktabs} 
\usepackage{cite}
\usepackage{amsmath}
\usepackage{amssymb}
\usepackage{amsfonts}
\usepackage{float}
\usepackage{graphicx}
\usepackage{hhline}
\usepackage{hyperref}
\usepackage{enumitem}
\usepackage{tikz}

\makeatletter
\renewcommand\footnoterule{%
  \kern-3\p@
  \hrule\@width 0.5\columnwidth
  \kern2.6\p@}
 \makeatother
 
\usepackage{fancyhdr,lipsum}
\fancypagestyle{firstpage}{
  \fancyhf{}
  \fancyhead[L]{Published as a journal paper at IEEE Design \& Test}
}
\begin{document}
\title{SSCNets: Robustifying DNNs using Secure Selective Convolutional Filters}
\author{\IEEEauthorblockN {Hassan Ali$^{1,*}$\thanks{*Hassan Ali and Faiq Khalid have equal contributions.}, 
         Faiq Khalid$^{2,*}$,
         Hammad Tariq$^1$,
         Muhammad Abdullah Hanif$^2$,\\
 		Rehan Ahmed$^1$,
 		Semeen Rehman$^2$, 
 		Muhammad Shafique$^2$}
  	\IEEEauthorblockA{$^1$National University of Sciences and Technology (NUST), Islamabad, Pakistan}	
  	\IEEEauthorblockA{$^2$Technische Universit\"at Wien (TU Wien), Vienna, Austria}
  	\IEEEauthorblockA{Email: \{rehan.ahmed, hali.msee17seecs, htariq.msee17seecs\}@seecs.edu.pk\\
  	 Email: \{faiq.khalid, muhammad.hanif, semeen.rehman, muhammad.shafique\}@tuwien.ac.at}\vspace{-20pt}
  	}
\maketitle
\thispagestyle{firstpage}
\begin{abstract}
	In this paper, we introduce a novel technique based on the Secure Selective Convolutional (SSC) techniques in the training loop that increases the robustness of a given DNN by allowing it to learn the data distribution based on the important edges in the input image. We validate our technique on Convolutional DNNs against the state-of-the-art attacks from the open-source Cleverhans library using the MNIST, the CIFAR-10, and the CIFAR-100 datasets. Our experimental results show that the attack success rate, as well as the imperceptibility of the adversarial images, can be significantly reduced by adding effective pre-processing functions, i.e., Sobel filtering.
\end{abstract}

\section{Introduction}\label{introduction}

Deep Neural Networks (DNNs) exhibit a strong dependence upon training data due to their underlying assumption that the inference data is sampled from the same distribution as the training dataset. This limitation of DNNs can be exploited by the adversaries to perform several security attacks (adversarial attacks), e.g., Fast Gradient Sign Method (FGSM)~\cite{goodfellow2014explaining}, Jacobian Saliency Map Attack (JSMA)~\cite{papernot2016limitations}, Basic Iterative Method (BIM), Carlini-Wagner Attacks (CW)~\cite{serban2018adversarial}, etc. 

To counter the adversarial attacks, several defense mechanisms have been proposed. One of the most prominent among these defenses is ``adversarial training" \cite{goodfellow2014explaining} which trains the network for known adversarial examples, hence, \textit{limits it to the known attacks.} A more versatile defense is required, which can defend against unknown adversarial attacks while maintaining the classification accuracy. As an alternative solution, pre-processing has emerged as a possible defense strategy against adversarial attacks \cite{khalid2019fademl}, e.g., noise filtering, feature squeezing, Gaussian data augmentation, MagNET, Generative Adversarial Network-based defenses\cite{serban2018adversarial}. \textit{However most of these defenses are either computationally expensive or become in-effective under white-box threat model.} Therefore, there is a dire need to develop efficient defense mechanisms that can effectively counter the attacks while maintaining the classification accuracy.

\begin{figure*}[!t]
	\centering
	\includegraphics[width=1\linewidth]{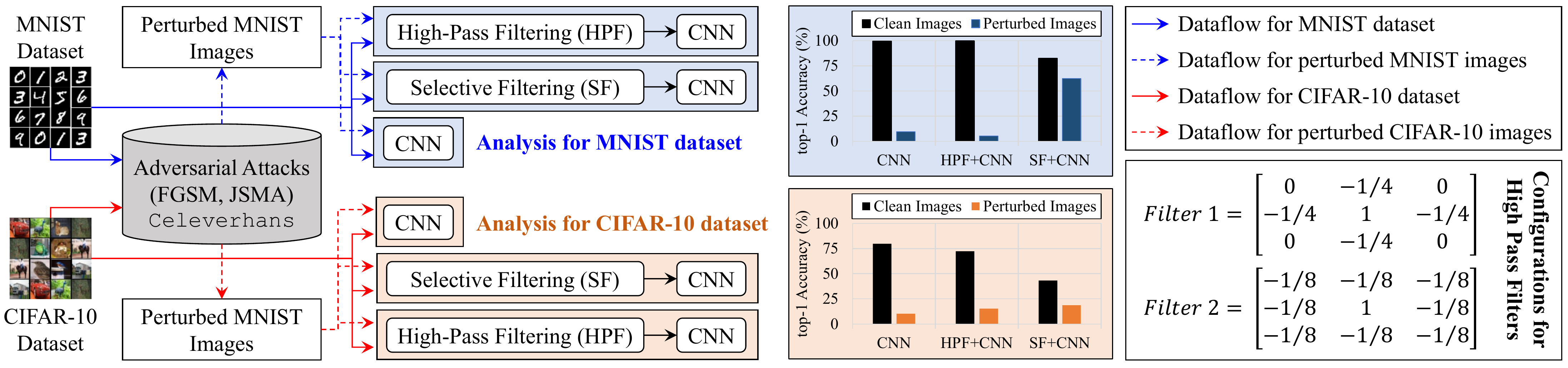}
	\caption{Experimental Setup for analyzing the effects of selective filtering on the robustness of CNNs. The graphs show the top-1 accuracy of the corresponding CNN architectures on clean images and perturbed images generated by FGSM attack. The perturbation value in each pixel is 0.3 for images from the MNIST dataset and 0.1 for images in the CIFAR-10 dataset.}
	\label{fig:Motivational}
\end{figure*}

\subsection{Motivational Analysis}\label{sec:MA}
Adversarial attacks affect the distribution of the pixel intensities of an image in such a way that it appears imperceptible in subjective and objective analysis~\cite{serban2018adversarial}. Therefore, training the DNN for the selective features may improve the classification accuracy of the DNNs. To analyze the effects of selective features, we implement the ``selective filtering'' at the input of the state-of-the-art CNNs\footnote{Conv(64, 8x8) - Conv(128, 6x6) - Conv(128, 5x5) - Dense(10) - Softmax()}, and perform the state-of-the-art FGSM attack from the open-source \texttt{cleverhans} library \cite{papernot2016cleverhans}, for both the MNIST and the CIFAR-10 datasets. For this analysis, we implement two types of high-pass filter (Filters 1 and 2 in Fig.~\ref{fig:Motivational}) configurations, i.e., the conventional filtering and the selective filtering, under the white-box setting. Figure~\ref{fig:Motivational} shows that the selective filtering shows a significant increase in the classification accuracy of the DNNs using both the MNIST and the CIFAR-10 dataset. \textit{However, the results identify that the defended CNN using the selective high-pass filtering shows a significant decrease in its accuracy on the unperturbed images}. These observations identify the following research challenges.
\begin{enumerate}
    \item How to increase the robustness (i.e., the resilience of the classifier towards perturbations in the input) of the CNN while keeping the accuracy on unperturbed images sufficiently high?
    \item How to select the different configurations of high-pass filterd and the appropriate threshold, $t_k$?
\end{enumerate}
\subsection{Novel Contributions}
To address the above-mentioned challenges, we retrain the DNN for important features of an input image by introducing the ``Selective Convolution (or Selective Filtering)" based pre-processing technique called as SSCNets, at the input of the DNN. These filters enhance the edges of the input image and suppressing the uniform based on a pre-defined threshold. The processed image is used to train the DNN, which is then used for prediction. Our novel contributions are as follows:
\begin{enumerate}
    \item We study and analyze the impact of different pre-processing noise filters on the robustness of the CNN under multiple configuration settings.
    \item Based on our analysis in Section~\ref{sec:MA}, we propose SSCNets, a novel and computationally efficient selective edge-triggered filtering method. This method integrates a preprocessing layer, Secure Selective Convolutional (SSC) Layer, with trained CNN to increase the robustness of the trained CNN.
\end{enumerate}

For illustration, we have analyzed SSCNets (SSC layer with a commonly used CNN architecture and SSC layer with VGG-16) for state-of-the-art adversarial attacks from an open-source \texttt{cleverhans} library.

\section{State-of-the-art Attacks and Defenses} \label{Background}
In this section, we provide a brief overview of the adversarial attacks and state-of-the-art defense mechanisms.
\subsection{Adversarial Attacks}
Adversarial attacks introduce specially crafted imperceptible perturbations in the input images to fool a DNN. The adversarial attacks can be categorized as follows: 

\subsubsection{Gradient-Based Attacks}
One of the most commonly used adversarial attacks is to exploit the gradient of the loss function with respect to the classification probability for a certain input sample. For example, FGSM~\cite{goodfellow2014explaining}, JSMA~\cite{papernot2016limitations}, BIM, C\&W and DeepFool, are some of the popular gradient-based attacks~\cite{serban2018adversarial}. However, computing the gradients based on the probability distributions, is computationally complex and expensive, which limits the applications of such iterative attacks. 

\subsubsection{Decision-based Attacks} 
Decision-based attacks utilize the class decision probability or decision labels of the DNN to generate the perturbation in the input image \cite{brendel2017decision}. However, such attacks are often very slow and require a large number of iterations to find the optimal adversarial perturbation.
\subsubsection{Score Based Attacks}
To exploit the probabilistic nature of DNNs and input samples, score-based attacks leverage the probabilistic behavior to estimate the gradients, i.e., Single-Pixel Attack and Local Search Attack \cite{rauber2017foolbox}.
\subsection{Defenses}
\begin{figure*}[!t]
	\centering
	\includegraphics[width=1\linewidth]{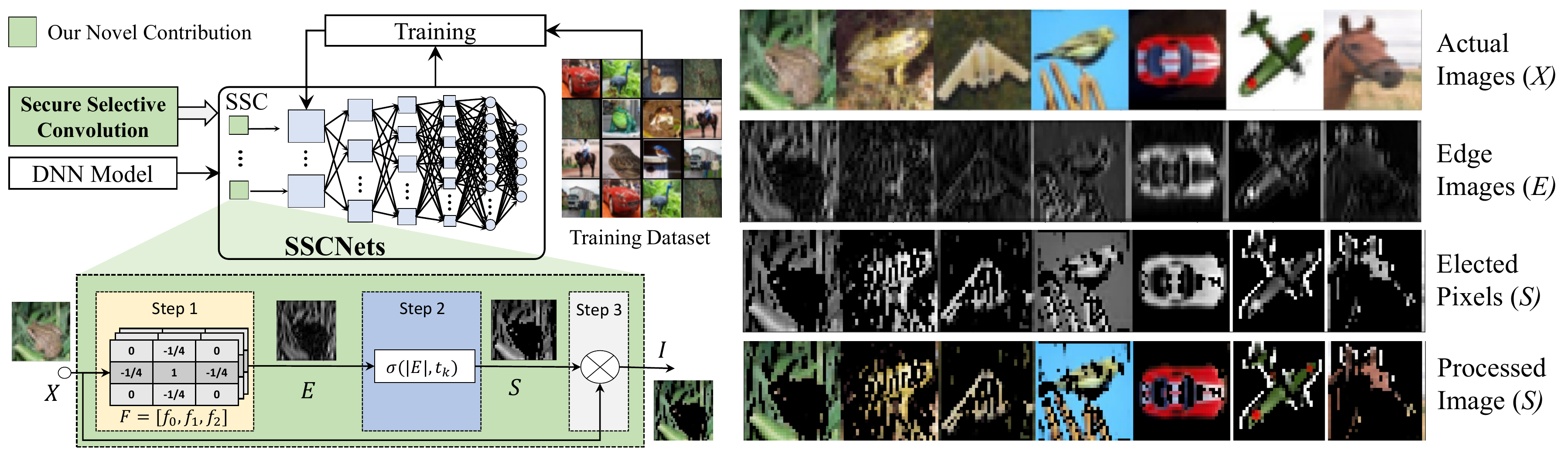}
	\caption{Secure Selective Convolution-based Defense (SSCNets) that integrates the Selective filtering with the CNN to increase its robustness with respect to adversarial attacks. Visualization examples to understand the intermediate steps of the SSCNets.}
	\label{fig:Methodology}
\end{figure*}

Adversarial (or perturbed) inputs computed by the inference attacks appear similar to the clean inputs for a human observer. This property of the adversarial inputs is known as \textit{imperceptibility}. Due to the imperceptible nature of inference attacks, they are one of the major security problems for CNN-based applications. To counter the adversarial attacks, several defense strategies have been proposed. These defenses can be categorized into three main categories.

\subsubsection{Gradient Masking}
Gradient masking, e.g., defensive distillation and thermometer encoding~\cite{serban2018adversarial}, masks the gradients of the DNN against gradient-based adversarial attacks. However, this approach can be compromised by black-box transfer attacks\footnote{Black-box transfer attack trains a substitute model to find the adversarial inputs for the substitute model and transfers these adversarial inputs to the target DNN} that can be used to estimate the gradients~\cite{carlini2017towards}.

\subsubsection{Adversarial Training}
Adversarial learning is one of the most commonly used defense strategies \cite{goodfellow2014explaining}. To perform the adversarial training, the defender has to first train a network for the clean inputs. Secondly, the defender computes the adversarial inputs for the trained network by performing an attack. The defender then retrains the network to make it resistant to the inference attacks. However, adversarial training is a supervised defense, i.e., it uses the known attack algorithms during the training process~\cite{rouhani2018deepfense}.
On the contrary, the proposed attack is an ``unsupervised defense'' because it does not consider the attack algorithm and adversarial examples to train the network. Hence, the attacks during the inference are ``unknown'' for the network~\cite{rouhani2018deepfense}.

\subsubsection{Pre-processing based Defenses}
Another defense strategy is to pre-process the input to reduce the effects of adversarial noise, e.g., MagNET, APE-GAN, defense-GAN, feature squeezing, and Gaussian data augmentation \cite{serban2018adversarial}. Feature Squeezing and Data Augmentation-based defenses~\cite{serban2018adversarial} are efficient defense mechanisms, but they fail under the white-box threat model~\cite{athalye2018obfuscated}. MagNET, APE-GAN, and defense-GAN transform the inputs into the manifold of training data using Generative Adversarial Networks (GANs). However, GANs~\cite{kurach2019large} are hard to train and are computationally very expensive as they utilize thousands of variables and multiplication operations. Therefore, it makes them ineffective for resource-constrained applications. On the other hand, SSCNets use only one convolutional filter for edge detection, one \texttt{sigmoid} activation layer, and one multiplication layer for pre-processing which makes it a suitable choice for real-time resource-constrained applications.

\section{Proposed Methodology} \label{Methodology}
To address the above-mentioned limitations, we propose to use the selective filtering as preprocessing (SSCNets) at the input of the given CNN.
We use two different types of filter configurations for our experiments. Each of these configurations is evaluated under different filter settings, as explained below.

\subsection{Conventional Filter Configuration}
It is the conventional filtering technique in which an input image \textit{X} is convoluted with the pre-defined filter \textit{F} and the output is fed into the subsequent CNN in the form of an image, \textit{I}. Mathematically, the input image \textit{X}, and the image \textit{I} that we feed to the subsequent CNN are related as follows:
\begin{equation}
    I = X \circledast F
\end{equation}
\begin{figure*}[!t]
	\centering
	\includegraphics[width=1\linewidth]{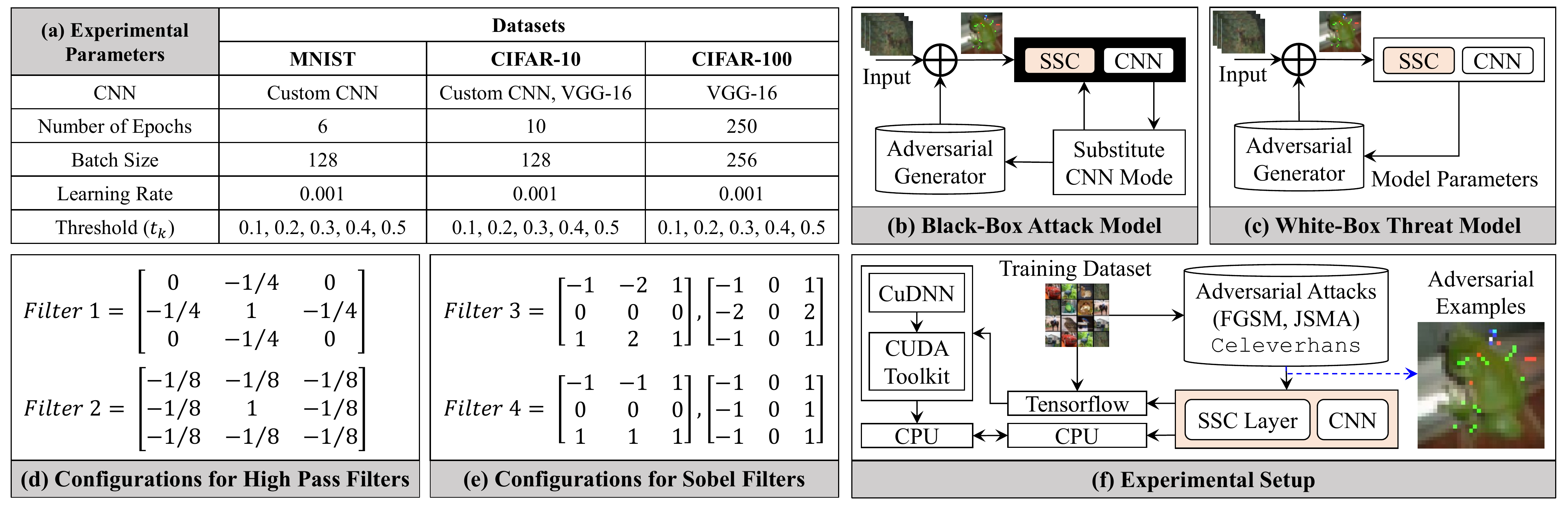}
	\caption{(a) Parameters used in this paper for experiments. Custom CNN: Conv(64, 8x8) - Conv(128, 6x6) - Conv(128, 5x5) - Dense(10) - Softmax() (b) The black-box attack model in which a substitute model is trained so that the parameters of the substitute model may be used in order to perform the attack. (c) The white-box attack model in which the parameters of the model itself are used to perform the attack. (d) Different configurations for high-pass filters used in this paper. (e) Different configurations for Sobel filters used in this paper. (f) The experimental setup used in this paper to evaluate the robustness of SSCNets.}
	\label{fig:Setup}
\end{figure*}
\subsection{Selective Filter Configuration}
In this section, we discuss a new methodology to enhance the robustness of a DNN against adversarial attacks by carefully identifying important pixels in the input image (see Figure~\ref{fig:Methodology}). We formally explain the complete procedure as follows.
\begin{enumerate}
    \item Let's assume a colored input image having three channels, i.e., $ X = [x^{(0)}, x^{(1)}, x^{(2)}]$ where, \(x^{(k)}\) denotes the \textit{k-th} channel of the input \textit{x}. We define the feature extraction mask, as follows,
    \begin{equation}
        F = [f^{(0)}, f^{(1)}, f^{(2)}]
    \end{equation}
    Where $f^{(k)}$ denotes the \textit{k-th} channel of the feature mask, i.e., in our case feature mask is $[f^{(0)}, f^{(1)}, f^{(2)}]$.
    
    \item To extract the edges of the input image, convolution is performed between the image \textit{X} and the feature extraction mask $F$ (which represents a high-pass filter) (For illustration purpose in Figure~\ref{fig:Methodology}, we use Laplacian Filter, a commonly used high-pass filter in Digital Image Processing),
    \begin{equation}
        E = X \circledast F
        \label{eq:3}
    \end{equation}
    In equation \ref{eq:3}, \textit{E} is a gray-scale image representing the “Edges” in \textit{X}.
    
    \item To select the strong edges, we set a threshold, $t_k$ (for our experiments, $t_k$ = 0.11). All the weaker edges (less than $t_k$) in image E, are zeroed out while the stronger edges are saturated to a maximum value which is 1 in our experiments. We use ``sigmoid'' function to perform strong edge selection, i.e. $S = \sigma (E, t_k)$, where $S$ varies from 0 to 1, with most of the values mapped to 0 and 1. The output $S$ is then multiplied with each channel of the input image $X$ to retain the color information.
    \begin{equation}
        I = X \times S
    \end{equation}
    \item Finally, the CNN is trained with the proposed preprocessing layer (SSC Layer).
\end{enumerate}
\section{Experimental Results} \label{Results}
\begin{figure*}[!t]
	\centering
	\includegraphics[width=1\linewidth]{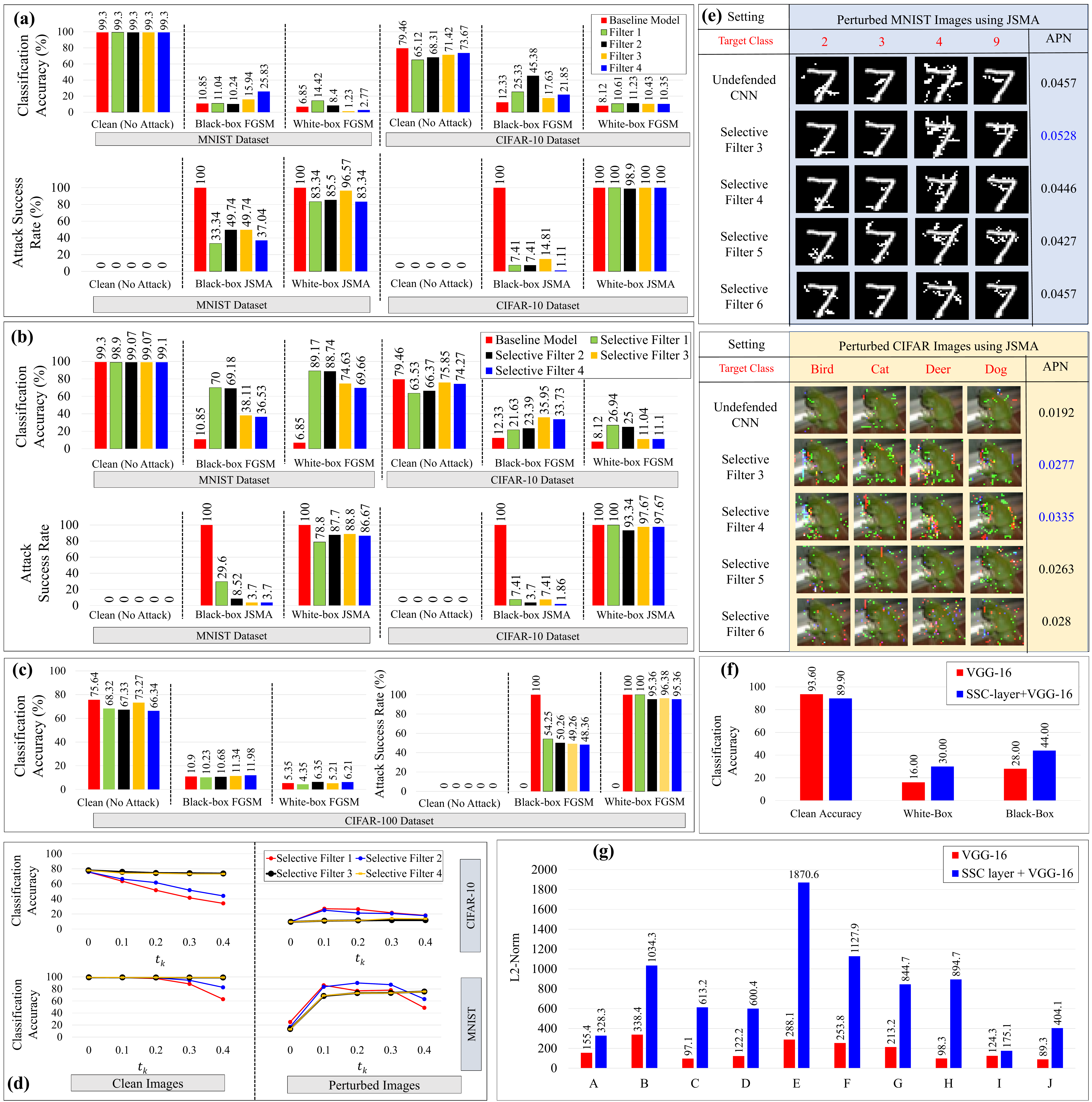}
	\caption{(a) The experimental results of non-selective convolutional techniques against both FGSM and JSMA attacks. (b) The results for secure selective convolution-based techniques (SSC) against the FGSM and the JSMA Attacks. (c) The experimental results for secure selective convolution-based techniques (SSC) against the FGSM and the JSMA Attacks for CIFAR-100 Dataset. Note, for more complex datasets like imagnet SSCNets requires identifying the proper configuration of the Sobel filters. (d) The effect of varying thresholds in SSCNet on the classification accuracy of clean and perturbed images generated from the FGSM attack with $\epsilon$ = 0.3 for MNIST and $\epsilon=0.1$ for CIFAR-10. (e) The comparison of adversarial examples for both the MNIST and the CIFAR-10 for different SSCNets configurations. APN represents the ``Average Perturbation Norm''. (f) The comparison of VGG-16 and Secure Selective Convolution (SSC)-based VGG-16 against the FGSM. (g) Comparison of VGG-16 with and without SSC based on the L2-Norm of adversarial images generated using the boundary attack for ten randomly chosen images from the CIFAR-10 dataset.}
	\label{fig:Results}
\end{figure*}

We implement multiple filter configurations as a pre-processing layer and integrate it with the state-of-the-art CNN from the Cleverhans library and evaluate the robustness of the CNN against a few of the adversarial attacks, i.e., FGSM, and JSMA (also from the Cleverhans library).
\subsection{Experimental Setup}
The experimental setup for our analysis is as follows (see Figure~\ref{fig:Setup}):

\begin{enumerate}
    \item \textbf{CNN:} Following the convention in the literature~\cite{carlini2017towards}\cite{papernot2016cleverhans}, we have used a commonly used CNN architecture commonly used to evaluate the robustness of various defense mechanisms, such as, Conv2D(64, 8x8) - Conv2D(128, 6x6) - Conv2D(128, 5x5)  - Dense(10) - Softmax() and VGG-16.
    
    \item \textbf{Datasets:} We have used the MNIST, the CIFAR-10, and the CIFAR-100 datasets for analyzing the SSCNets. To evaluate the robustness against ``unknown'' attacks, we trained the classifier on clean images only. The experimental configurations and experimental setup are shown in Figure \ref{fig:Setup}.
    
    \item \textbf{Attacks:} In this analysis, we have performed state-of-the-art FGSM and JSMA attacks.  
    
    \textbf{Epsilon ($\epsilon$):} It represents the noise added in the pixel on a scale of 0 to 1.
    
    \item \textbf{Attack Models:} We assume following attack scenarios:
    \begin{enumerate}
        \item \textbf{White-Box Scenario:} The attacker has complete information about the defense mechanism, the defense parameters, the network architecture, and the weights of the classifier (see Figure~\ref{fig:Setup}(c)).
        \item \textbf{Black-Box Scenario:} The attacker has no information about the defense mechanism or the network architecture and the weights of the classifier. In this case, the attacker performs the transfer attack using a substitute CNN model (see Figure~\ref{fig:Setup}(b)).
    \end{enumerate}
    \item \textbf{Metrics:} To evaluate the robustness of the model, we use two commonly used metrics, i.e., attack success rate\footnote{The ratio of adversarial inputs which cause the classifier to make incorrect predictions to the total number of adversarial inputs generated for the purpose. For a given constraint of perturbation and number of iterations, the lower the attack success rate, the more robust the classifier.}, for the JSMA attack, and classification accuracy\footnote{The accuracy of the classifier on the inputs perturbed using the Adversarial Attacks. The greater the classification accuracy, the more robust the classifier.} of the model on Perturbed Images, for FGSM attack. Moreover, to compute the perceptibility, we use one of the most commonly used metrics, i.e., perturbation norm\footnote{Perturbation norm is defined as \\ $L2-Norm=\sum_{all pixels} (Clean Image-Perturbed Image)^2$}. 
    \item \textbf{Filters:} The filter settings for our experiments are given in Figure~\ref{fig:Setup}(d). We use these settings because unlike other commonly used filters in digital image processing (e.g., median filters, max filters, and min filters), these filters are differentiable and hence do not mask the gradients of the classifier, which can lead to misleading results~\cite{athalye2018obfuscated}.
\end{enumerate}

\subsection{Experimental Analysis}
We analyzed the impact of both selective and non-selective filters, as the pre-processing filtering layer in the proposed SSCNets under white-box and black-box settings. 

\subsubsection{\textbf{Impact of Non-Selective Filtering}} 
To analyze the impact of non-selective filtering, we used four different kinds of filter settings, as shown in Figure \ref{fig:Setup}(d) (Filter 1, 2, 3, 4). By analyzing the results in Figure \ref{fig:Results}(a), we make the following observations:
\begin{enumerate}
    \item In the black-box setting, a significant decrease in the attack success rate for both the FGSM and the JSMA attacks has been observed. The success rate of JSMA decreases rapidly as compared to the FGSM attack. Hence, it can be concluded that the transferability of the FGSM attack is more powerful than the JSMA Attack.
    \item In the white-box threat model, the non-selective pre-processing filters did not show any positive impact on the robustness of the CNN. This is because, under the white-box scenario, these attacks incorporate the effects of pre-processing filters in the optimization algorithms.
\end{enumerate}
\subsubsection{\textbf{Impact of Selective Edge-triggered Filtering}}
Figure \ref{fig:Results}(b) reports the impact of the selective edge-triggered filtering on the attack success rates of the FGSM and the JSMA for four different filter settings, as shown in Figure~\ref{fig:Setup}(d) (Selective Filter 3, 4, 5, 6). We analyzed the results and made the following observations:

\begin{enumerate}
    \item For the black-box threat model, we observed a significant decrease in the success rate of both FGSM and JSMA attacks.
    \item For the white-box threat model, we observed that selective filtering effectively decreases the effectiveness of the FGSM attack. However, against the JSMA attacks, selective filtering does not give promising results. This is because of the iterative nature of the JSMA attack which makes it more powerful compared to the FGSM attack. 
\end{enumerate}

To show the applicability of SSCNets on larger networks, we have also evaluated it on the VGG-16 network. The experimental results show that VGG-16 without the SSC layer provides 93.7\% accuracy on CIFAR-10 after 250 epochs. After introducing the SSC layer in VGG-16, the accuracy drops to 89.9\% (see Figure~\ref{fig:Results}(e)). However, the accuracy of VGG-16 with the SSC layer on adversarial examples increases significantly, i.e., from 16 \% to 30\%, which is almost two times, for the white-box scenario and from 28\% to 44\% for the black-box scenario.

\subsubsection{\textbf{Impact of $t_k$ in Selective Edge-triggered Filtering}}
Figure \ref{fig:Results}(c) shows the effects of SSCNets threshold $t_k$ on the classification accuracy of the CNN with secure selective convolution. In can be observed that with the increase in threshold $t_k$ (i.e., from 0.0 to 0.4), the robustness offered by the SSCNets increases for both the MNIST, CIFAR-10, and CIFAR-100 datasets. This is because the large values of $t_k$ result in the CNN being trained on more significant edges. \textit{However, this increased robustness comes at the cost of the decreased classification accuracy on unperturbed (clean) images.} Moreover, for larger values of threshold $t_k$ (i.e., $>>0.1$) the robustness of the CNN decreases exponentially. This is because the impact on classification becomes more prominent. Based on the analysis, we observed that the optimum value for the threshold is between 0.1 and 0.2; therefore, in our analysis, we choose it to be $t_k=0.11$.


\subsubsection{\textbf{Perceptibly of Adversarial Attacks}}
Figure \ref{fig:Results}(d) shows the comparison of the adversarial images generated using the JSMA attack for different configurations of SSCNets. From the analysis, it can be observed that the adversarial images against a defended CNN are more perceptible than those generated against an undefended CNN (especially for CIFAR-10 and CIFAR-100). This is because that the CNN is trained on the most relevant features of the training dataset that makes it more sensitive to the changes near the edges. Consequently, the perturbation around the edges introduced by the JSMA attack becomes more perceptible. We have also observed that the perturbation norms of adversarial images for MNIST dataset are comparable under different filter settings. This may be because MNIST images already have very strong edge information.
\subsubsection{\textbf{Boundary Attack}}
Decision-based attacks guarantee an adversarial example if a sufficient number of queries are allowed~\cite{cheng2018query}. Therefore, instead of comparing the success rate of decision-based attacks, researchers usually predefine the number of iterations and consider the L2-norm of the adversarial images for comparison~\cite{cheng2018query}. Following the same approach, we randomly choose ten images from CIFAR-10 and try to fool VGG-16 and SSC-VGG-16 (SSC-layer + VGG-16) networks by performing the state-of-the-art boundary attack~\cite{brendel2017decision} with a maximum number of queries set to 25000. Boundary attack achieves a 100\% success rate against both architectures. Figure~\ref{fig:Results}(f) reports the L2-norm of the adversarial images found by the boundary attack against VGG-16 and SSC-VGG-16 for the ten chosen images. \textit{We found that the average value of L2-norm of adversarial images for SSC-VGG-16 architecture is 789.3325 which is 443\% greater than that for VGG-16, i.e.,l 178.00565.}

\section{Conclusion}\label{Conclusion}
In this paper, we propose to integrate the Secure Selective Convolutional (SSC) layer with CNN to increase its robustness against the adversarial attacks. Though it slightly reduces the classification accuracy of the CNN on the clean inputs, it significantly improves the overall robustness against different adversarial noise, and also increases the perceptibility of adversarial noise. We experimentally demonstrated it against state-of-the-art FGSM and JSMA attacks using the CIFAR-10, CIFAR-100, and the MNIST datasets.
\section*{Acknowledgement}
This work was partially supported by the Erasmus+ International Credit Mobility (KA107). 
\bibliographystyle{IEEEtran.bst}
\bibliography{SSCNets.bbl}
\vspace{5mm}

\begin{itemize}
    \item \textbf{Hassan Ali} is a post-graduate student at NUST, Pakistan. He is interested in embedded systems, machine Learning, machine learning security and artificial intelligence. 
    
    \item \textbf{Faiq Khalid} is currently pursuing a Ph.D. degree in hardware security at TU Wien, Austria. His research interests include hardware security, IoT and machine learning. He has received the Quaid-e-Azam Gold Medal for his academic achievements, and the prestigious Richard Newton Fellowship at DAC 2018. 
    
    \item \textbf{Hammad Ali Tariq} is a post-graduate student at NUST, Pakistan. He is interested in machine learning and machine learning security.
    
    \item \textbf{Muhammad Abdullah Hanif} is currently a University Assistant at TU Wien, Austria. His research interests include brain-inspired computing, machine learning, approximate computing, computer architecture, energy-efficient design, robust computing, and emerging technologies. He is a recipient of the President's Gold Medal for his outstanding academic performance during his M.S. degree.
    
    \item \textbf{Rehan Ahmed} is an assistant professor at NUST, Pakistan. His current research interests include low-power FPGA architectures/CAD, and FPGA-based system designing. He has a Ph.D. in electrical and computer engineering from the University of British Columbia, Canada (2015). 
    
    \item \textbf{Semeen Rehman} is an assistant professor at TU Wien, Austria. Her current research interests include cross-layer reliability modeling and optimization, approximate computing, and embedded systems. She has a Ph.D. in computer science from Karlsruhe Institute of Technology, Germany (2015). 
    
    \item \textbf{Muhammad Shafique} is currently a Full Professor with the Institute of Computer Engineering, Department of Informatics, Technische University Wien (TU Wien), Vienna, Austria, where he is directing the group on Computer Architecture and Robust, Energy-Efficient Technologies. His research interests are in computer architecture, energy-efficient systems, robust computing, hardware security, brain-inspired computing, emerging technologies, and embedded systems.. 
\end{itemize}

\end{document}